\documentclass[10pt,twocolumn,letterpaper]{article}

\usepackage{cvpr}              

\usepackage[accsupp]{axessibility}  

\usepackage[normalem]{ulem}
\usepackage{graphicx}
\usepackage{amsmath}
\usepackage{amssymb}
\usepackage{booktabs}
\usepackage{xcolor}
\usepackage{comment}
\usepackage{enumitem}

\newcommand{\benchmark}{WireSegHR}

\usepackage[pagebackref,breaklinks,colorlinks]{hyperref}

\usepackage[capitalize]{cleveref}
\crefname{section}{Sec.}{Secs.}
\Crefname{section}{Section}{Sections}
\Crefname{table}{Table}{Tables}
\crefname{table}{Tab.}{Tabs.}

\def\cvprPaperID{699} 
\def\confName{CVPR}
\def\confYear{2023}

\begin{document}

\title{Automatic High Resolution Wire Segmentation and Removal}

\author{Mang Tik Chiu$^{1,2}$, Xuaner Zhang$^2$, Zijun Wei$^{2}$, Yuqian Zhou$^2$, Eli Shechtman$^2$, \\Connelly Barnes$^{2}$, Zhe Lin$^2$, Florian Kainz$^2$, Sohrab Amirghodsi$^2$, Humphrey Shi$^{1,3}$ 
\\
{\small $^1$UIUC, $^2$Adobe, $^3$University of Oregon}\\
{\small \textbf{\url{https://github.com/adobe-research/auto-wire-removal}}}}

\twocolumn[{%
\renewcommand\twocolumn[1][]{#1}%
\maketitle
\begin{center}
    \centering
    \captionsetup{type=figure}
    \includegraphics[width=\textwidth]{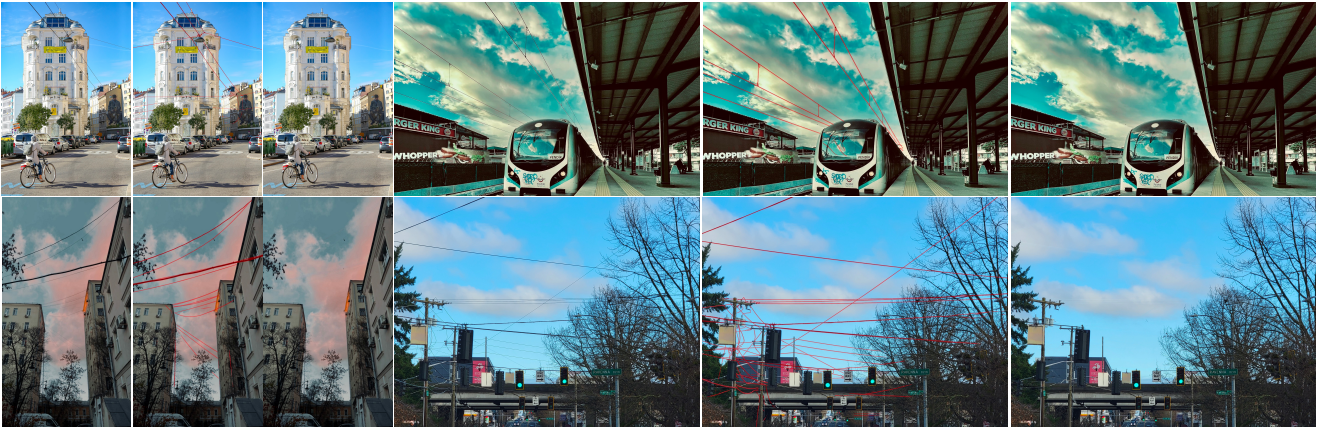}
    \captionof{figure}{We present an automatic high-resolution wire segmentation and removal pipeline. Each triad shows the high-resolution input image, our automatic wire segmentation result masked in red, and our full-resolution wire removal result. The visual quality of these photographs is greatly improved with our fully-automated wire clean-up system.}
    \label{fig:demo}
\end{center}%
}]


\begin{abstract}
\vspace{-5.5mm}
Wires and powerlines are common visual distractions that often undermine the aesthetics of photographs. The manual process of precisely segmenting and removing them is extremely tedious and may take up hours, especially on high-resolution photos where wires may span the entire space. In this paper, we present an automatic wire clean-up system that eases the process of wire segmentation and removal/inpainting to within a few seconds. 
We observe several unique challenges: wires are thin, lengthy, and sparse. These are rare properties of subjects that common segmentation tasks cannot handle, especially in high-resolution images. 
We thus propose a two-stage method that leverages both global and local contexts to accurately segment wires in high-resolution images efficiently, and a tile-based inpainting strategy to remove the wires given our predicted segmentation masks. We also introduce the first wire segmentation benchmark dataset, \benchmark. Finally, we demonstrate quantitatively and qualitatively that our wire clean-up system enables fully automated wire removal with great generalization to various wire appearances.

\end{abstract}


\section{Introduction}


Oftentimes wire-like objects such as powerlines and cables can cross the full width of an image and ruin an otherwise beautiful composition. Removing these ``distractors'' is thus an essential step in photo retouching to improve the visual quality of a photograph. Conventionally, removing a wire-like object requires two steps: 1) segmenting out the wire-like object, and 2) removing the selected wire and inpainting with plausible contents. Both steps, if done manually, are extremely tedious and error-prone, especially for high-resolution photographs that may take photographers up to hours to reach a high-quality retouching result.

In this paper, we explore a fully-automated wire segmentation and inpainting solution for wire-like object segmentation and removal with tailored model architecture and data processing.
For simplicity, we use \textit{wire} to refer to all wire-like objects, including powerlines, cables, supporting/connecting wires, and objects with wire-like shapes.

Wire semantic segmentation has a seemingly similar problem setup with generic semantic segmentation tasks; they both take in a high-resolution image and generate dense predictions at a pixel level.
However, wire semantic segmentation bears a number of unique challenges. First, wires are commonly long and thin, oftentimes spanning the entire image yet having a diameter of only a handful of pixels. A few examples are shown in Figure~\ref{fig:motivation}. This prevents us from getting a precise mask based on regions of interest. Second, the input images can have arbitrarily high resolution up to 10k$\times$10k pixels for photographic retouching applications. Downsampling such high-resolution images can easily cause the thin wire structures to disappear. This poses a trade-off between preserving image size for inference quality and run-time efficiency.
Third, while wires have regular parabolic shapes, they are often partially occluded and can reappear at arbitrary image location, thus not continuous.
(e.g.~\cite{lanedet,swiftlane}).

To account for these challenges, we propose a system for automatic wire semantic segmentation and removal. For segmentation, we design a two-stage coarse-to-fine model that leverages both pixel-level details in local patches and global semantics from the full image content, and runs efficiently at inference time. For inpainting, we adopt an efficient network architecture~\cite{suvorov2022resolution}, which enables us to use a tile-based approach to handle arbitrary high resolution. We design a training strategy to enforce color consistency between the inpainted region and the original image.
We also present the first benchmark dataset, \benchmark, for wire semantic segmentation tasks, where we collect and annotate high-resolution images with diverse scene contents and wire appearances.
We provide analyses and baseline comparisons to justify our design choices, which include data collection, augmentation, and our two-stage model design. Together, these design choices help us overcome the unique challenges of accurately segmenting wires.
Our contributions are as follows: 

\begin{itemize}[noitemsep]
  \item \textbf{Wire segmentation model:} We propose a two-stage model for wire semantic segmentation that leverages global context and local information to predict accurate wire masks at high resolution. We design an inference pipeline that can efficiently handle ultra-high resolution images.
  \item \textbf{Wire inpainting strategy:} We design a tile-based inpainting strategy and tailor the inpainting method for our wire removal task given our segmentation results.
  \item \textbf{\benchmark, a benchmark dataset:} We collect a wire segmentation benchmark dataset that consists of high resolution images, with diversity in wire shapes and scene contents. We also release the manual annotations that have been carefully curated to serve as ground truths. Besides, we also propose a benchmark dataset to evaluate inpainting quality. 
\end{itemize}

\section{Related Work} \label{sec:related}

\begin{figure}[t!]
\centering
\includegraphics[width=1.0\linewidth]{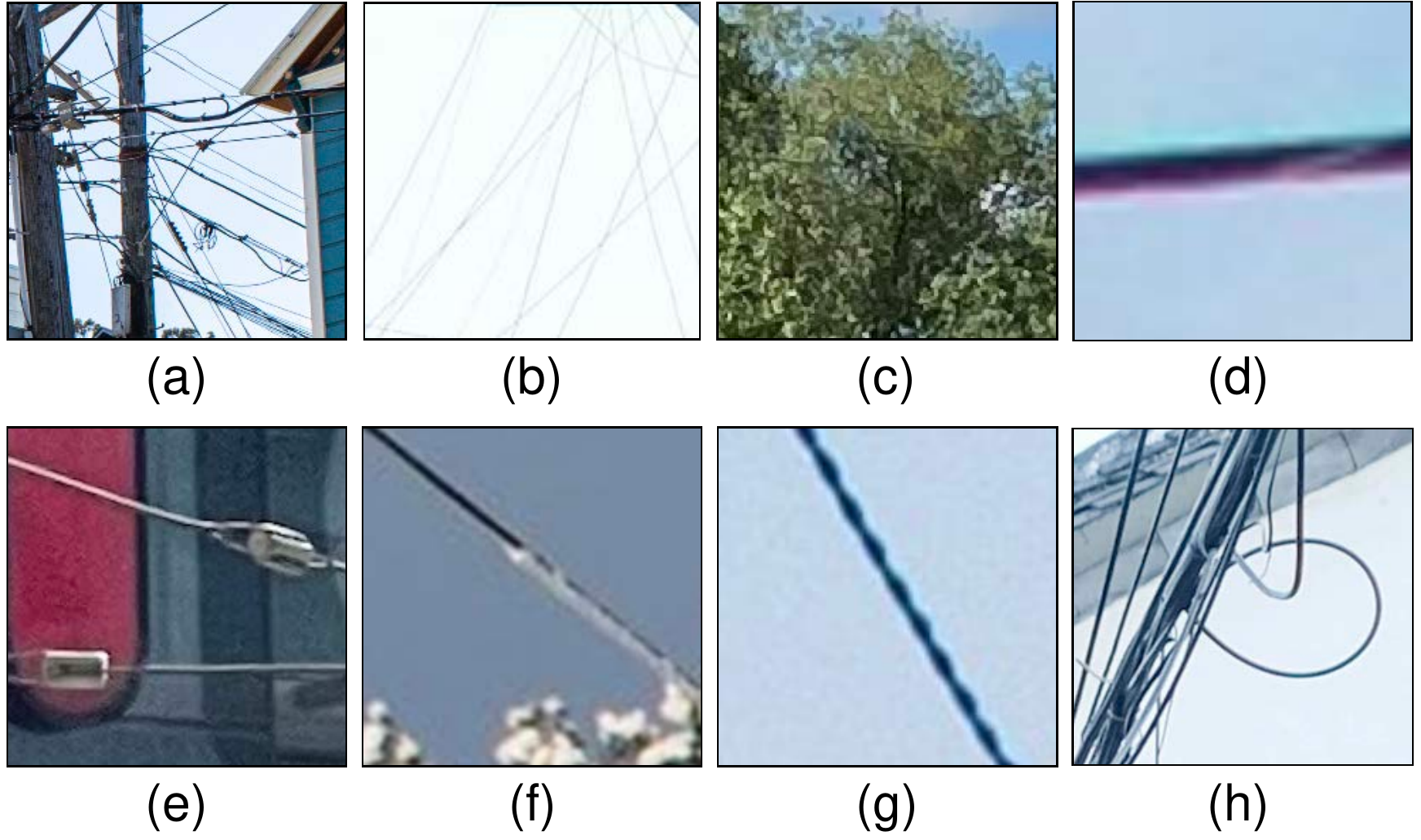}
\caption{\textbf{Challenges of wire segmentation.} Wires have a diverse set of appearances. Challenges include but are not limited to (a) structural complexity, (b) visibility and thickness, (c) partial occlusion by other objects, (d) camera aberration artifacts, and variations in (e) object attachment, (f) color, (g) width and (h) shape.
}
\vspace{-5.5mm}
\label{fig:motivation}
\end{figure}



\paragraph{Semantic segmentation}
Semantic segmentation has been actively researched over the past decade. For example, the DeepLab series~\cite{deeplab, deeplabv3, deeplabv3p} has been one of the most widely used set of semantic segmentation methods. They leverage dilated convolutions to capture long-range pixel correlations. Similarly, CCNet~\cite{ccnet} attend to non-local regions via a two-step axial attention mechanism. PSPNet~\cite{pspnet} use multi-scale pooling to extract high-resolution features.

Recently, the self-attention mechanism~\cite{attention} has gained increasing popularity. 
Transformer-based models for semantic segmentation~\cite{dpt, setr, swin, segformer, hassani2022neighborhood, hassani2022dilated, jain2021semask, jain2022oneformer} significantly outperform convolution-based networks since the attention modules benefit from their global receptive fields~\cite{segformer}, which let the models attend to objects that span across larger portions of the feature map.

While these above methods work well in common object semantic segmentation, when applied to our task of wire segmentation in high-resolution images, they either drop significantly in segmentation quality or require long inference times. We show in Section~\ref{sec:results} that directly applying these methods to our task yields undesirable results.

\vspace{-5mm}
\paragraph{High-resolution image segmentation}
Segmentation in high-resolution images involves additional design considerations. It is computationally infeasible to perform inference on the full-resolution image with a deep network. As a result, to maximally preserve image details within the available computation resources, many methods employ a global-local inference pipeline. For instance, GLNet~\cite{glnet} simultaneously predict a coarse segmentation map on the downsampled image and a fine segmentation map on local patches at the original resolution, then fuse them to produce the final prediction. 
MagNet~\cite{magnet} is a recent method that proposes to iteratively predict and refine coarse segmentation maps at multiple scales using a single feature extractor and multiple lightweight refinement modules. CascadePSP~\cite{cascadepsp} train a standalone class-agnostic model to refine predictions at a higher resolution from a pretrained segmentation model. ISDNet~\cite{isdnet} propose to use an extremely lightweight subnetwork to take in the entire full-resolution image. However, the subnetwork is limited in capacity and thus segmentation quality. We share the same idea with these past works on using a coarse-to-fine approach for wire segmentation, but modify the architecture and data processing to tailor to wires.

\vspace{-5mm}
\paragraph{Wire/Curve segmentation}
While few works tackle wire segmentation in high-resolution images, there are prior works that handle similar objects. For example, Transmission Line Detection (TLD) is an actively researched area in aerial imagery for drone applications. Convolutional neural networks are used~\cite{ttpla, pldu, cable_inst, lsnet} to segment overhanging power cables in outdoor scenes. However, wire patterns in TLD datasets are relatively consistent in appearance and shape -- evenly spaced and only spanning locally. In contrast, we handle more generic wires seen in regular photographic contents, where the wire appearance has much higher variety. 

Some other topics are loosely related to our task. Lane detection~\cite{lanedet,swiftlane,structurelane} aims to segment lanes for autonomous driving applications. These methods benefit from simple line parameterization (e.g., as two end-points), and strong positional priors. In contrast, as shown in Figure~\ref{fig:motivation}, wires vary drastically in shapes and sizes in our task, thus making them difficult to parameterize.

\vspace{-5mm}
\paragraph{High-Resolution Image Inpainting}
Image inpainting has been well-explored using patch synthesis-based methods \cite{barnes2009patchmatch, wexler2007space, darabi2012image, kaspar2015self} or deep neural networks \cite{contextencoder, globallocal, partialconv, contextual, yu2019free, xu2022image}. Zhao \textit{et al.} leveraged the powerful image sysnthesis ability of StyleGAN2 \cite{karras2020analyzing} and proposed CoModGAN \cite{comodgan} to push the image generation quality to a newer level, and was followed by \cite{zheng2022cm, jain2022keys}. Most of these deep models cannot be applied to inpainting tasks at high-resolution images. The latest diffusion-based inpainting model like DALLE-2 \cite{dalle}, LDM \cite{rombach2022high}, and StableDiffusion etc. also suffer from long inference time and low output resolution. ProFill \cite{zeng2020high} was first proposed to address high resolution inpainting via a guided super resolution module. HiFill \cite{hifill} utilized a contextual residual aggregation module and the resolution can be up to 8K. LaMa \cite{suvorov2022resolution} applied the fourier convoluational residual blocks to make the propagation of image structures well. LaMa was trained on only $256 \times 256$ images, but can be used for images up to 2K with high quality. Recently, Zhang \textit{et al.} \cite{supercaf} proposed to use guided PatchMatch for any-resolution inpainting and extended the deep inpainting results from LaMa to modern camera resolution. The textures are better reused, while the structure and line completion at high-resolution can still be challenging. In this paper, we aim at removing wires from high resolution photos. The problem can become easier if we run inpainting in a local manner since wires are usually thin and long. Therefore, we propose to revisit LaMa for wire removal, and run the inference in a tile-based fashion. 

\section{Dataset Collection and \benchmark} \label{sec:dataset}

\subsection{Image Source and Annotations}

Our definition of wires include electrical wires/cables, power lines, supporting/connecting wires, and any wire-like object that resemble a wire structure.
We collect high-resolution images with wires from two sources: 80\% of the images are from photo sharing platforms (Flickr, Pixabay, etc.), and 20\% of the images are captured with different cameras (DSLRs and smartphones) in multiple countries on our own. For the internet images, we collect 400K candidate images by keyword-searching. Then, we remove duplicates and images where wires are the only subjects. We then curate the final 6K images that cover sufficient scene diversity like city, street, rural area and landscape.


\begin{figure}[t!]
    \centering
    \captionsetup{type=figure}
    \includegraphics[width=\linewidth]{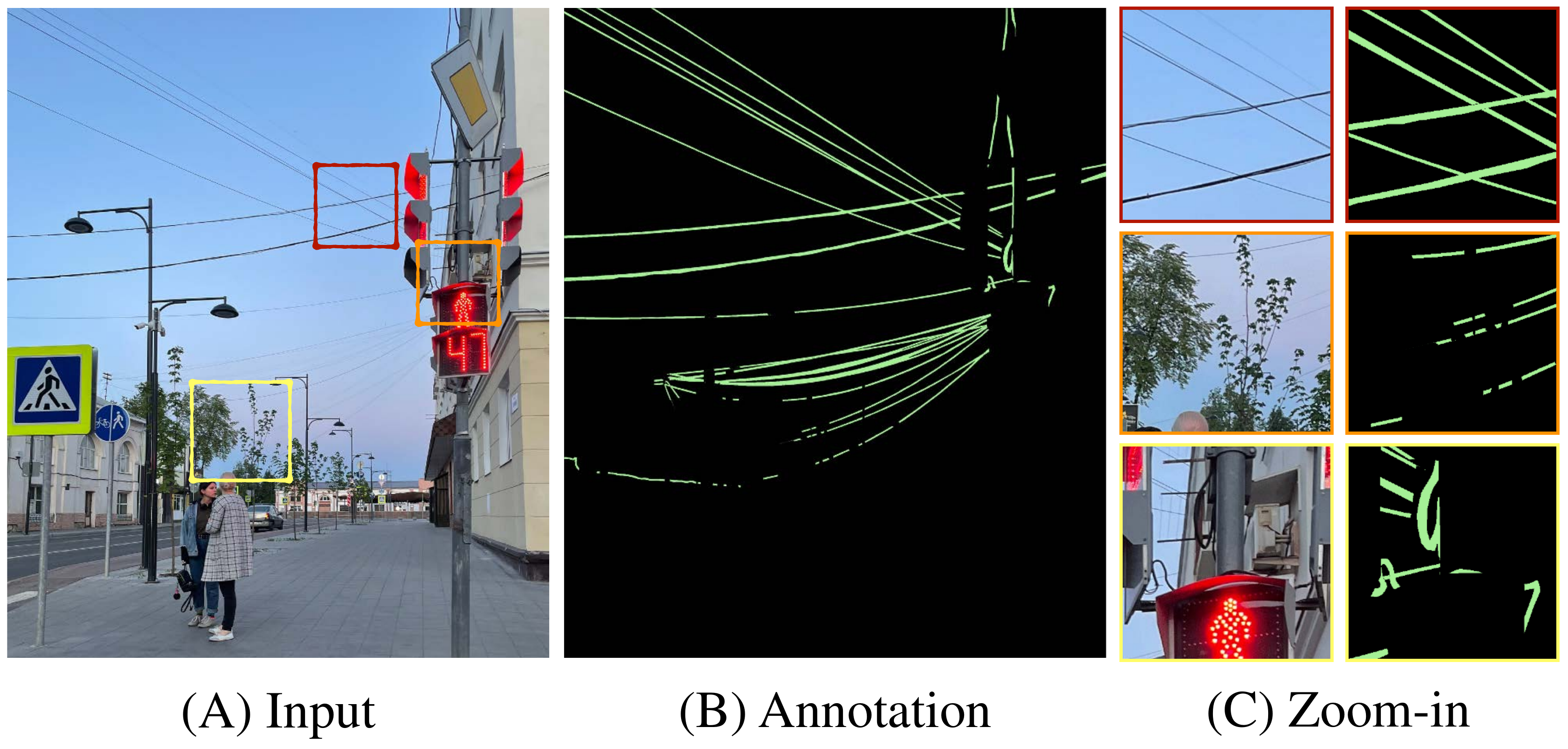}\\
    \vspace{-1mm}
    \captionof{figure}{\textbf{Wire Annotation Example.} An example wire annotation in our dataset. Our annotation (B) is accurate in different wire thicknesses (\textcolor{red}{red}), variations in wire shapes (\textcolor{orange}{orange}) and accurate wire occlusions (\textcolor{yellow}{yellow}).}
\vspace{-4mm}
    \label{fig:annotation}
\end{figure}
\begin{figure*}[hbt!]
\centering
\includegraphics[width=1.0\textwidth]{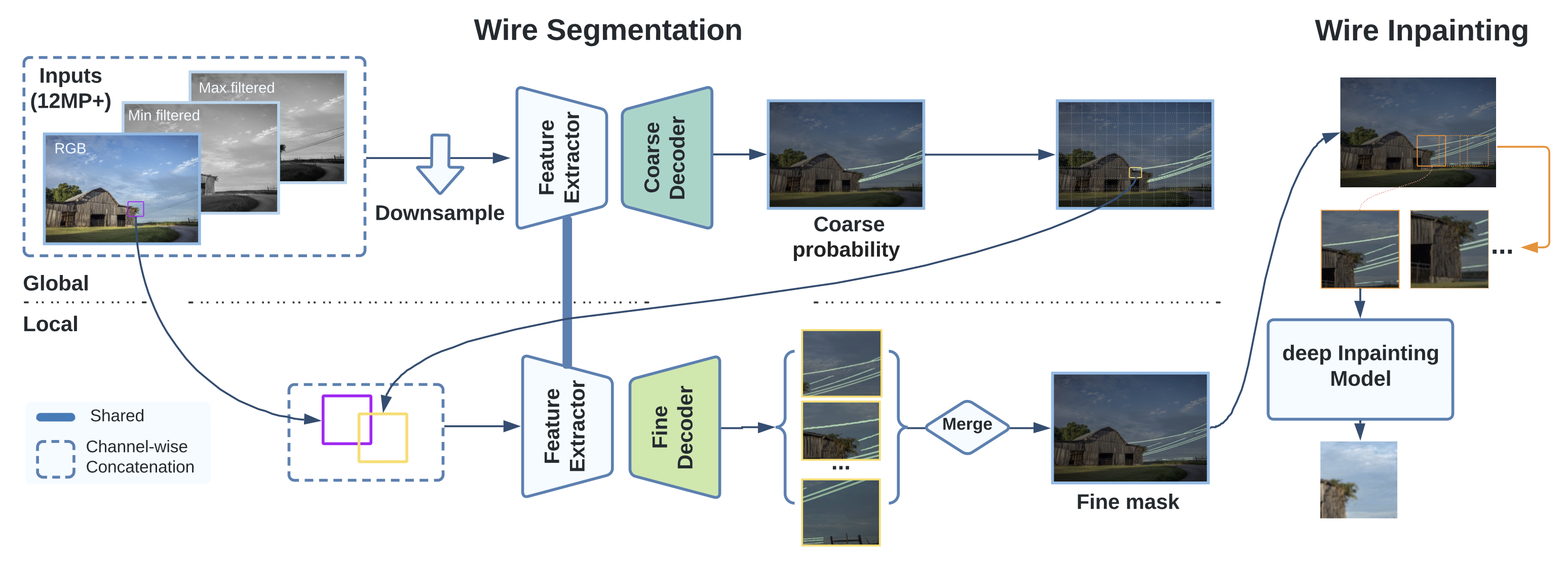}
\vspace{-2mm}
\caption{\textbf{Our wire removal system}. A system overview of our wire segmentation and removal for high resolution images. Input is concatenated with min- and max-filtered luminance channels. The downsampled input is fed into the coarse module to obtain the global probability. In the local stage, original-resolution patches are concatenated with the global probability map to obtain the local logit map. After a segmentation mask is predicted, we adopt LaMa architecture and use a tile-based approach to achieve wire removal. See Section~\ref{sec:segmentation},~\ref{sec:inpainting} for details.}
\vspace{-1mm}
\label{fig:pipeline}
\end{figure*}
Our wire annotation process contains two rounds. In the first round, annotators draw detailed masks over wires at full-resolution. The annotated masks enclose the main wire body and the boundary, oftentimes including a gradient falloff due to aliasing or defocus.
The boundary region annotation is crucial so as to avoid residual artifacts during wire removal.
In the second round, quality assurance is carried out to re-annotate unsatisfactory annotations. We show an example of our high-quality wire annotations in Figure~\ref{fig:annotation}. 

\subsection{Dataset Statistics}

In Table~\ref{table:stats}, we list the statistics of our dataset and compare them with existing wire-like datasets. Our dataset is the first wire dataset that contains high-resolution photographic images. The dataset is randomly split into 5000 training, 500 validation, and 500 testing images. We release 420 copyright-free test images with annotations.


\begin{table}[h!]
\centering
\resizebox{\linewidth}{!}{
    \renewcommand{\arraystretch}{1.1}
    \begin{tabular}{r|cccc}
    \hline
    
    Dataset & \begin{tabular}[x]{@{}c@{}}\# Wire\\Images\end{tabular}  & \begin{tabular}[x]{@{}c@{}}Min.\\Image Size\end{tabular}  &\begin{tabular}[x]{@{}c@{}}Max.\\Image Size\end{tabular}  &\begin{tabular}[x]{@{}c@{}}Median\\Image Size\end{tabular} \\ \hline
    Powerline~\cite{powerlinedataset} & 2000 & 128$\times$128 & 128$\times$128 & 128$\times$128\\
    PLDU~\cite{pldu} & 573 & 540$\times$360 & 540$\times$360 & 540$\times$360 \\
    PLDM~\cite{pldu} & 287 & 540$\times$360 & 540$\times$360 & 540$\times$360 \\
    TTPLA~\cite{ttpla} & 1100 & 3840$\times$2160 & 3840$\times$2160 & 3840$\times$2160\\ \hline
    \textbf{Ours} & 6000 & 360$\times$240 & 15904$\times$10608 & 5040$\times$3360 \\ \hline

\end{tabular}
}
\vspace{-2mm}
\caption{Statistics of our wire dataset compared to others.}
\vspace{-5mm}
\label{table:stats}
\end{table}





\section{High-Resolution Wire Segmentation}
\label{sec:segmentation}
Wires appear visually different from common objects -- being thin, long, sparse and oftentimes partially occluded. 
We find the following two design choices crucial to building an effective wire segmentation system: 1) having a two stage framework so that coarse prediction from global context guides precise segmentation from local patches and 2) maximally preserving and augmenting image features and annotations of wires throughout the pipeline.  

\subsection{The Two-stage Coarse to Fine Model}
Figure \ref{fig:pipeline} shows the two-stage segmentation pipeline. It consists of a coarse and a fine module, which share an encoder $E$ and have their own decoder $D_C$ and $D_F$. Intuitively, the coarse module aims to capture the global contextual information from the entire image and highlight the image regions possibly containing wires. Conditioned on the predictions from the coarse module, the fine module achieves high-resolution wire segmentation by only looking at local patches likely containing wires.

Given a high-resolution image $I_\textrm{glo}$, we first bilinearly downsample it to $I_\textrm{glo}^{ds}$ with a fixed size $p\times p$ and feed it into the coarse module. The module predicts the global probability map $P_\textrm{glo} = \textrm{SoftMax}(D_C(E(I_\textrm{glo}^{ds})))$ containing the activation of the wire regions.

For each patch $I_\textrm{loc}$ of size $p \times p$ cropped from the full-resolution image $I_\textrm{glo}$, and the corresponding conditional probability map $P_\textrm{con}$ cropped from $P_\textrm{glo}$, we predict the local probability $P_\textrm{loc} = \textrm{SoftMax}(D_F(E(I_\textrm{loc}, P_\textrm{con})))$. 
Note that $E$ is shared between the coarse and the fine module, thus it should take inputs with the same number of channels. Therefore, for the coarse module, we concatenate an additional zero channel with the input image to make the channel number consistent.


We apply Cross Entropy (CE) loss to both the global $P_\textrm{glo}$ and local probability map $P_\textrm{loc}$, comparing with their ground truth annotations $G_\textrm{glo}$ and $G_\textrm{loc}$.

\vspace{-2mm}
\begin{equation}
\begin{aligned}
    \mathcal{L}_\textrm{glo} &= CE (P_\textrm{glo}, G_\textrm{glo}) \\
    \mathcal{L}_\textrm{loc} &= CE (P_\textrm{loc}, G_\textrm{loc}) \\
\end{aligned}
\end{equation}
The final loss $\mathcal{L}$ is the sum of the two:

\vspace{-3mm}
\begin{equation}
\begin{aligned}
    \mathcal{L} &= \mathcal{L}_ \textrm{glo} + \lambda \mathcal{L}_ \textrm{loc},
\end{aligned}
\end{equation}

where we 
set $\lambda = 1$ for training. 
Similar to Focal loss~\cite{focal} and Online Hard Example Mining~\cite{ohem}, we balance the wire and background samples in the training set by selecting patches that contain at least 1\% of wire pixels. 


To perform inference, we first feed the downsampled image to the coarse module, which is the same as training. 
Local inference is done by running a sliding window over the entire image, where the patch is sampled only when there is at least some percentage of wire pixels (determined by~$\alpha$). This brings two advantages: First, we save computation time in regions where there are no wires. Second, the local fine module can leverage the information from the global branch for better inference quality.


\subsection{Wire Feature Preservation}
\label{sec:wire_feature_preserve}
As wires are thin and sparse, applying downsampling to the input images may make the wire features vanish entirely. To mitigate this challenge, we propose a simple feature augmentation technique by taking the min and max pixel luminance values of the input image over a local window. Either the local min or the max value makes the wire pixels more visually apparent. In practice, we concatenate the min- and max-filtered luminance channels to the RGB image and condition map, resulting in 6 total channels as input. We name this component MinMax.

Besides feature augmentations, we also adapt the architecture to maximally preserve the sparse wire annotations.
We propose to use ``overprediction" and achieve this by using max-pool downsampling on the coarse labels during training, which preserves activation throughout the coarse branch. We name this component MaxPool. We provide ablation studies for these components in Section~\ref{sec:results}.

\vspace{-5mm}
\section{High-Resolution Wire Inpainting}
\label{sec:inpainting}
Given a full-resolution wire segmentation mask estimated by our wire segmentation model, we propose an inpainting pipeline to remove and fill in the wire regions. Our approach addresses two major challenges in wire inpainting. First, recent state-of-the-art deep inpainting methods do not handle arbitrary resolution images, which is critical for high-resolution wire removal. Second, deep inpainting methods often suffer from color inconsistency when the background has uniform (or slowly varying) colors. This issue is particularly significant for wires, as they are often in front of uniform backgrounds, such as the sky or building facades. The commonly used reconstruction loss, such as L1, is not sensitive to color inconsistency, which further exacerbates this issue.

We thus revisit the efficient deep inpainting method LaMa \cite{suvorov2022resolution}. Compared with other inpainting models, LaMa has two major advantages. First, it contains the Fourier convolutional layers which enables an efficient and high-quality structural completion. This helps complete building facades and other man-made structures with fewer artifacts. Second, its high inference efficiency makes a tile-based inference approach possible for high resolution images. 

To address color inconsistency, we propose a novel ``onion-peel" color adjustment module. Specifically, we compute the mean of the RGB channels within the onion-peel regions $M_o = D(M, d) - M$ of the wire mask $M$, where $D$ is the binary dilation operator, and $d$ is the kernel size. The color difference for each channel $c \in {R, G, B}$ becomes $\textrm{Bias}_c = \mathbb{E}[M_o (x_c - y_c)]$, where $x$ is the input image, and $y$ is the output from the inpainting network. The final output of the inpainting model is: $\hat{y_c} = y_c + \textrm{Bias}_{c}$. Loss functions are then applied to $\hat{y_c}$ to achieve color consistency while compositing the final result $y_{out} = (1 - M) \odot x + M \odot \hat{y}$. 

\section{Experiments} \label{sec:results}

\subsection{Implementation Details}
\paragraph{Wire Segmentation Network.}
We experiment with ResNet-50~\cite{resnet} and MixTransformer-B2~\cite{segformer} as our shared feature extractor. We expand the input RGB channel to six channels by concatenating the conditional probability map, min- and max-filtered luminance channels. For the min and max filtering, we use a fixed 6x6 kernel. We use separate decoders for the coarse and fine modules, denoted as $D_C$ and $D_F$ respectively.

We use the MLP decoder proposed in~\cite{segformer} for the MixTransformer segmentation model, and the ASPP decoder in~\cite{deeplabv3p} for our ResNet-50 segmentation model. In both the segmentation and inpainting modules, we take the per-pixel average of the predicted probability when merging overlapping patches. To crop $P_\mathrm{con}$ from $P_\mathrm{glo}$, we upsample the predicted $P_\mathrm{glo}$ to the original resolution, then crop the predicted regions according to the sliding window position.

To train the segmentation module, we downsample the image $I_\mathrm{glo}$ to $p\times p$ to obtain $I_\mathrm{glo}^\mathrm{ds}$. From $I_\mathrm{glo}$, we randomly crop one $p\times p$ patch $I_\mathrm{loc}$ that contains at least 1\% wire pixels. This gives a pair of $I_\mathrm{glo}^\mathrm{ds}$ and $I_\mathrm{loc}$ to compute the losses. During inference, $I_\mathrm{glo}^\mathrm{ds}$ is obtained in the same way as training, while multiple $I_\mathrm{loc}$ are obtained via a sliding window sampled only when the proportion of wire pixels is above $\alpha$. All feature extractors are pretrained on ImageNet.

We train our model on 5000 training images. The model is trained for 80k iterations with a batch size of 4. We set patch size $p = 512$ during training. 
For all ResNet models, we use SGD with a learning rate of 0.01, a momentum of 0.9 and weight decay of 0.0005. For MixTransformer models, we use AdamW~\cite{adamw} with a learning rate of 0.0002 and weight decay of 0.0001. Our training follows the ``poly" learning rate schedule with a power of 0.9. During inference, we set both the global image size and local patch size $p$ to 1024. Unless otherwise specified, we set the percentage for local refinement to $1\%$ ($\alpha=0.01$).


\vspace{-4mm}
\paragraph{Wire Inpainting Network.}
We adopt LaMa \cite{suvorov2022resolution} for wire inpainting by finetuning on an augmented wire dataset. 
To prepare the wire training set, we randomly crop ten $680\times680$ patches from the non-wire regions of each image in our training partition. In total, we have 50K more training images in addition to 
the 8M
Places2 \cite{zhou2017places} dataset, and increase its sampling rate 
by $10\times$
to balance the dataset. We also use all the ground truth segmentation maps in our training set to sample wire-like masks. During training, we start from Big-LaMa weights, and train the model on $512\times 512$ patches. We also prepare a synthetic wire inpainting quality evaluation dataset, containing 1000 images at $512\times 512$ with synthetic wire masks.
While running inference on full-resolution images, we apply a tile-based approach, by fixing the window size at $512\times 512$ with an $32$-pixel overlap. 

\subsection{Wire Segmentation Evaluation}
\paragraph{Quantitative Evaluation}


We compare with several widely-used object semantic segmentation and high-resolution semantic segmentation models. Specifically, we train DeepLabv3+~\cite{deeplabv3p} with ResNet-50~\cite{resnet} backbone under two settings: global and local. In the global setting, the original images are resized to 1024$\times$1024. In the local setting, we randomly crop 1024$\times$1024 patches from the original images. We train our models on 4 Nvidia V100 GPUs and test them on a single V100 GPU. For high-resolution semantic segmentation models, we compare with CascadePSP~\cite{cascadepsp}, MagNet~\cite{magnet} and ISDNet~\cite{isdnet}. We describe the training details of these works in the supplement.

We present the results of in Table~\ref{table:results} tested on \benchmark. We report wire IoU, F1-score, precision and recall for quantitative evaluation. We also report wire IoUs for images at three scales, small (0 -- 3000$\times$3000), medium (3000$\times$3000 -- 6000$\times$6000) and large (6000$\times$6000+), which are useful for analyzing model characteristics. Finally, we report average, minimum and maximum inference times on \benchmark.

\begin{table*}[t!]
\vspace{2mm}
\centering
\resizebox{0.98\textwidth}{!}{
    \renewcommand{\arraystretch}{1.1}
    \addtolength{\tabcolsep}{-2pt}
    \begin{tabular}{r|c|ccc|ccc|ccc}
    \hline
    Model & \begin{tabular}[x]{@{}c@{}}Wire\\IoU\end{tabular} & F1 & Precision & Recall & \begin{tabular}[x]{@{}c@{}}IoU\\(Small)\end{tabular} & \begin{tabular}[x]{@{}c@{}}IoU\\(Medium)\end{tabular} & \begin{tabular}[x]{@{}c@{}}IoU\\(Large)\end{tabular} & \begin{tabular}[x]{@{}c@{}}Avg. Time\\(s/img)\end{tabular} & \begin{tabular}[x]{@{}c@{}}Min. Time\\(s/img)\end{tabular} & \begin{tabular}[x]{@{}c@{}}Max. Time\\(s/img)\end{tabular}\\ \hline\hline 
    DeepLabv3+ (Global)~\cite{deeplabv3p} & 37.77 & 54.83 & 69.68 & 45.20 & 51.62 & 38.89 & 31.89 & 0.22 & 0.07 & 0.78 \\
    DeepLabv3+ (Local)~\cite{deeplabv3p} & 48.66 & 65.46 & 68.13 & 63.0 & 60.23 & 51.44 & 40.17 & 3.27 & 0.05 & 16.59 \\ \hline
    CascadePSP (Pretrained)~\cite{cascadepsp} & 20.44 & 33.94 & 62.19 & 23.34 & 33.64 & 21.80 & 13.78 & 2.32 & 0.37 & 36.79 \\
    CascadePSP (Retrained)~\cite{cascadepsp} & 26.85 & 42.33 & 52.44 & 35.49 & 48.22 & 28.97 & 15.80 & 2.25 & 0.37 & 25.37 \\
    MagNet~\cite{magnet} & 33.71 & 50.42 & 87.69 & 35.38 & 43.59 & 32.67 & 34.48 & 3.89 & 0.54 & 17.97\\
    MagNet-Fast~\cite{magnet} & 37.87 & 54.94 & 67.98 & 46.09 & 46.75 & 35.88 & 41.42 & 1.36 & 0.55 & 5.33 \\
    ISDNet (R-18)~\cite{isdnet} & 46.52 & 63.50 & 77.56 & 53.75 & 55.09 & 47.15 & 43.34 & 0.29 & 0.12 & 0.86\\
    ISDNet (MiT-b2)~\cite{isdnet} & 47.90 & 64.77 & 77.38 & 55.70 & 54.48 & 46.77 & 49.51 & 0.26 & 0.13 & 1.02 \\ \hline
    Ours (R-50) & 47.75 & 64.64 & 74.86 & 56.87 & 60.68 & 50.19 & 38.19 & 1.24 & 0.13 & 4.67 \\ 
    Ours (MiT-b2) & 60.83 & 75.65 & 83.62 & 69.06 & 63.52 & 59.83 & 62.93 & 0.82 & 0.07 & 3.36 \\ \hline
\end{tabular}
\addtolength{\tabcolsep}{2pt}
}
\vspace{-1mm}
\caption{Performances of common semantic segmentation and recent high-resolution semantic segmentation models on our dataset. We find that our dataset poses many challenges that high-resolution segmentation models fail to tackle effectively.
}
\label{table:results}
\end{table*}


As shown in Table~\ref{table:results}, while the global model runs fast, it has lower wire IoUs. In contrast, the local model produces high-quality predictions, but requires a very long inference time.
Meanwhile, although CascadePSP is a class-agnostic refinement model designed for high-resolution segmentation refinement, it primarily targets common objects and does not generalize to wires. 
For MagNet, its refinement module only takes in probability maps without image features, thus failing to refine when the input prediction is inaccurate. 
Among these works, ISDNet is relatively effective and efficient at wire prediction. 
However, their shallow network design trades off capacity for efficiency, limiting the performance of wire segmentation that is thin and sparse. 

Compared to the methods above, our model achieves the best trade-off between accuracy and memory consumption. By leveraging the fact that wires are sparse and thin, our pipeline captures both global and local features more efficiently, thus saving a lot of computation while maintaining high segmentation quality.

\begin{figure*}[t!]
\centering
\includegraphics[width=1.0\linewidth]{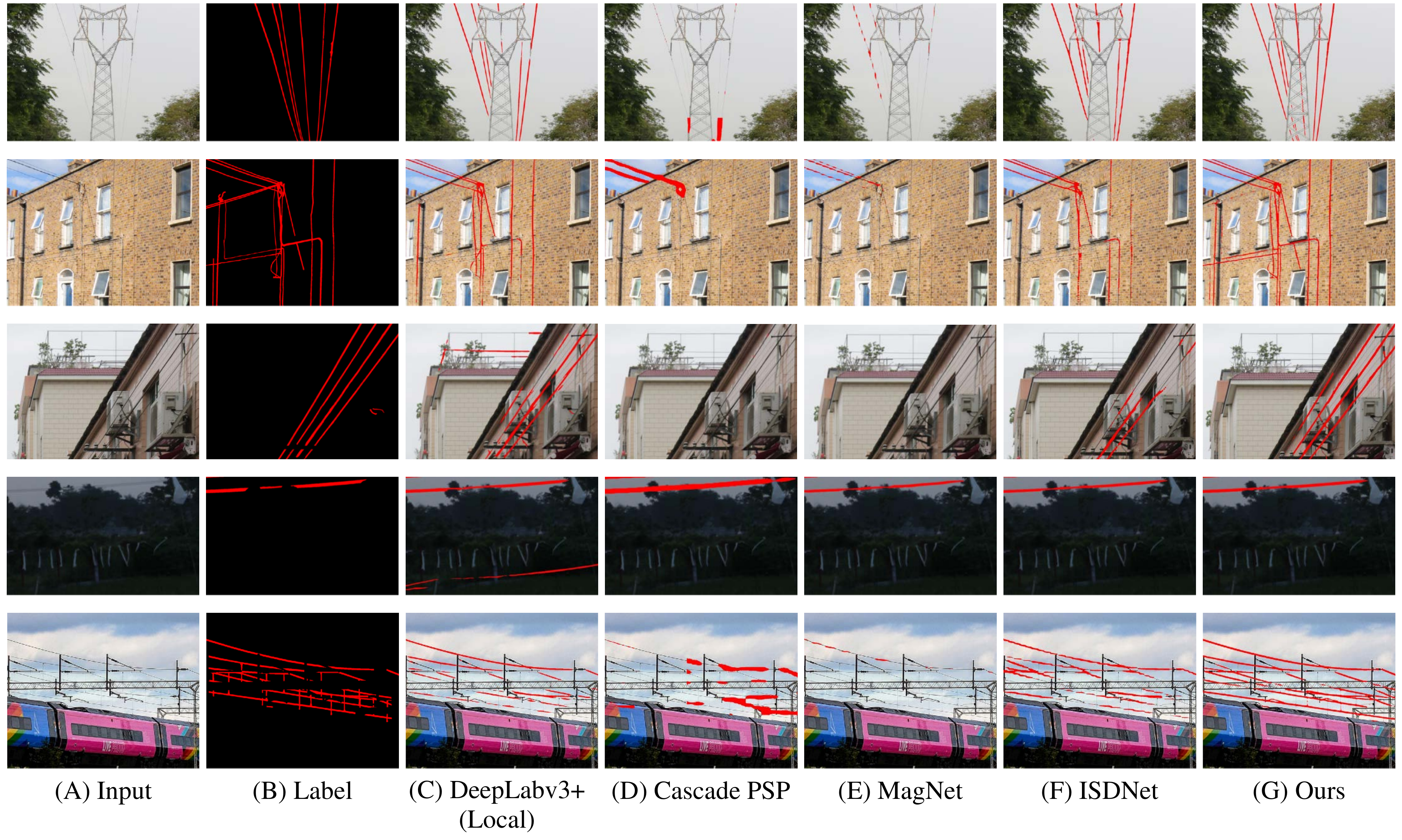}
\vspace{-8mm}
\caption{
\textbf{Qualitative comparison of several semantic segmentation models.} 
A common object semantic segmentation model (DeepLabv3+) either fails to find thin wires or overpredicts due to lack of global context. On the other hand, CascadePSP and MagNet, being refinement-based models, cannot work well on wires when the predictions are inaccurate or missing. While ISDNet can capture many thin wires regions, it cannot produce a high-quality prediction. In contrast, our model is able to both capture accurate wire regions and produce fine wire masks, and maintain low inference time.
\vspace{-5mm}
}
\label{fig:visual}
\end{figure*}

\begin{figure*}[t!]
\centering
\vspace{1mm}
\includegraphics[width=1.0\linewidth]{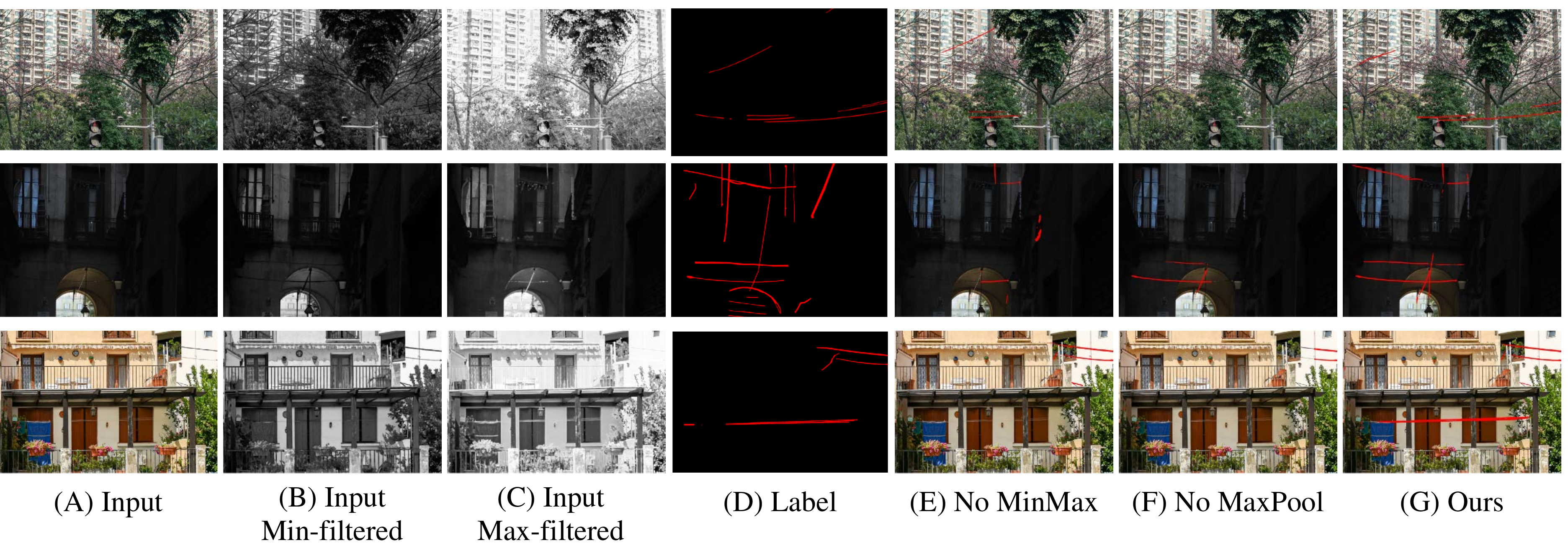}
\vspace{-7mm}
\caption{
\textbf{Qualitative comparison of our model components.} MinMax enhances wire image features when they are too subtle to see in RGB, while MaxPool encourages aggressive predictions in the coarse branch. Both components enable the model to pick up more regions for the final wire mask prediction.
}
\label{fig:visual}
\end{figure*}


\vspace{-4.5mm}
\paragraph{Qualitative Evaluation}
We provide visual comparisons of segmentation models in Figure~\ref{fig:visual}. We show the ``local'' DeepLabv3+ model as it consistently outperforms its ``global'' variant given that ``local'' predicts wire masks in a sliding-window manner at the original image resolution. As a trade-off, without global context, the model suffers from over-prediction. CascadePSP is designed to refine common object masks given a coarse input mask, thus fails to produce satisfactory results when the input is inaccurate or incomplete. Similarly, the refinement module of MagNet does not handle inaccurate wire predictions. ISDNet performs the best among related methods, but the quality is still unsatisfactory as it uses a lightweight model with limited capacity. Compared to all these methods, our model captures both global context and local details, thus producing more accurate mask predictions.
\vspace{-4mm}
\paragraph{Ablation Studies}






In Table~\ref{table:component_ablation}, we report wire IoUs after removing each component in our model, including MinMax, MaxPool, and Coarse condition concatenation. We find that all components play a significant role for accurate wire prediction, particularly in large images. Both MinMax and MaxPool are effective in encouraging prediction, which is shown by the drop in recall without either component, also shown in Figure~\ref{fig:visual}. Coarse condition, as described in Section~\ref{sec:segmentation}, is crucial in providing global context to the local network, without which the wire IoU drops significantly.

Table~\ref{table:thresholds} shows the wire IoUs and inference speed of our two-stage model as $\alpha$ changes. We observe a consistent decrease in performance as $\alpha$ increases. On the other hand, setting $\alpha$ to 0.01 barely decreases IoU, while significantly boosting inference speed, which means the coarse network is effectively activated at wire regions.







\subsection{Wire Inpainting Evaluation}
\begin{figure}[h!]
\centering
\captionsetup{type=figure}
\includegraphics[width=1.\linewidth]{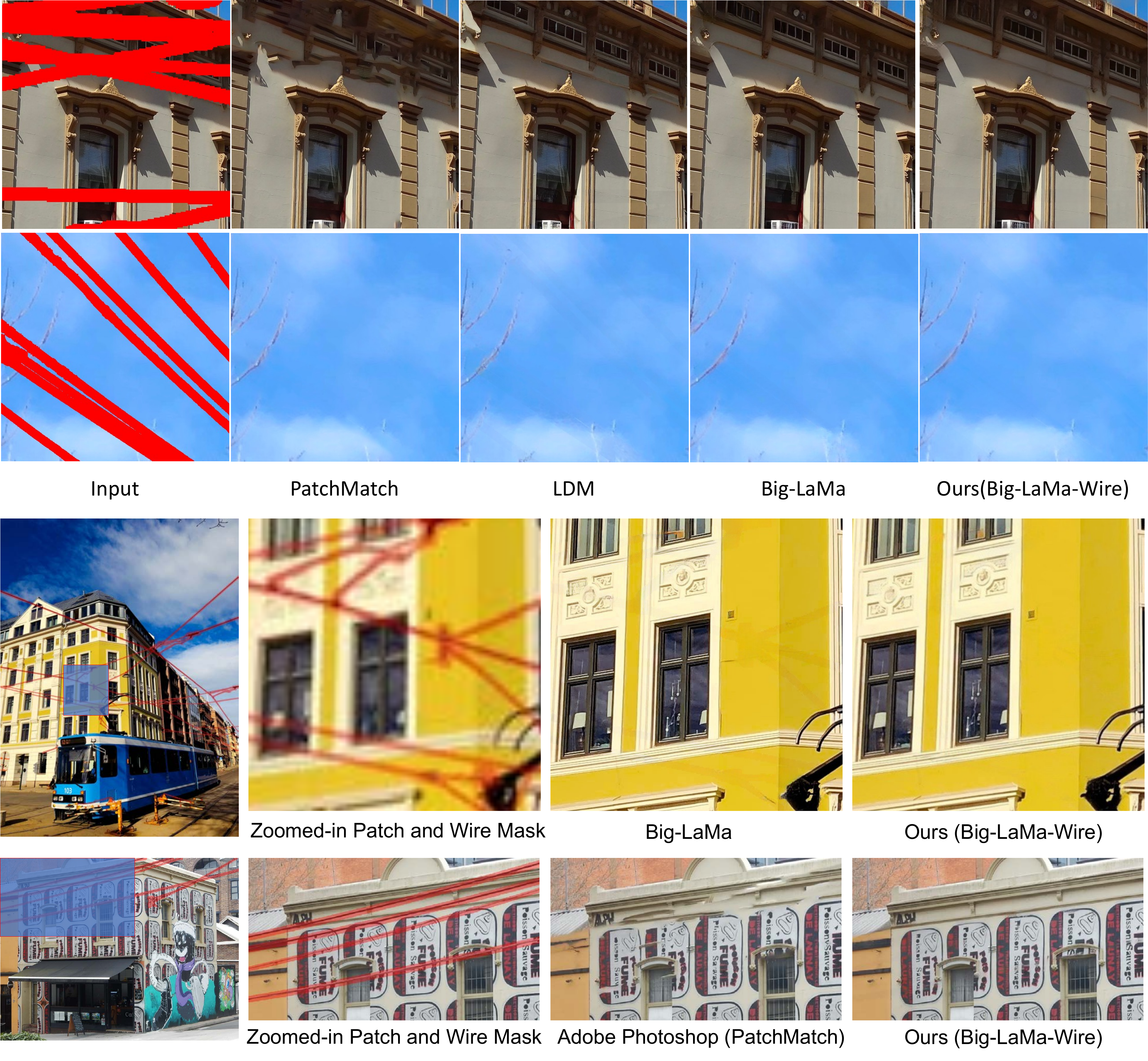}
\vspace{-6mm}
\captionof{figure}{\textbf{Inpainting Comparison}. Our model performs well on complicated structure completion and color consistency, especially on building facades and sky regions containing plain and uniform color. 
\vspace{-3mm}
}
\label{fig:wire_inp}
\end{figure}

\vspace{-1mm}
We evaluate our wire inpainting model using the synthetic dataset. Results are shown in Table~\ref{exp:wire_inp}. Our model structure is highly related to LaMa~\cite{suvorov2022resolution}. The difference is the training data and the proposed color adjustment module to address color inconsistency. We also compare our methods with PatchMatch \cite{barnes2009patchmatch} based on patch synthesis, DeepFillv2~\cite{yu2019free} based on Contextual Attention, CMGAN~\cite{zheng2022cm} and FcF~\cite{jain2022keys} based on StyleGAN2~\cite{karras2020analyzing} and LDM~\cite{rombach2022high} based on Diffusion. Inference speed is measured on a single A100-80G GPU. Visual results on synthetic and real images are shown in Figure \ref{fig:wire_inp}. PatchMatch, as a traditional patch synthesis method, produces consistent color and texture that leads to high PSNR. However, it performs severely worse on complicated structural completion. StyleGAN-based CMGAN and FcF are both too heavy for wires that are thin and sparse. Besides, diffusion-based models like LDM tends to generate arbitrary objects and patterns. DeepFill and the official Big-LaMa both have severe color inconsistency issue, especially in the sky region. Our model has a balanced quality and efficiency, and performs well on structural completion and color consistency. 
Note that we use a tile-based method at inference time.
The reason the tile-based strategy can be employed is due to the wire characteristics: sparse, thin and lengthy. More high-resolution inpainting results are in the supplementary materials.




\section{Discussion}

\vspace{-1.5mm}
\subsection{Comparison with Google Pixel 6}
\vspace{-1.5mm}
Recently, Google Pixel 6~\cite{Pixel6as49} announced the ``Magic Eraser'' photo feature that automatically detects and removes distractors. Note that this is a product feature and is not specifically designed for wires, and thus is hardly comparable with our method. We compare against this feature by uploading the images to Google Photos and applying ``Magic Eraser'' without manual intervention. We find that ``Magic Eraser'' performs well on wires with clear background, but it suffers from thin wires that are hardly visible and wires with complicated background. We show two examples in the supplementary material.

\subsection{Failure cases}
\vspace*{-2.5mm}

While our proposed wire segmentation model produces high-quality masks in most situations, there are still some challenging cases that our model cannot resolve. In particular, wires that are heavily blended in with surrounding structures/background, or wires under extreme lighting conditions are challenging to segment accurately. We show several examples in the supplementary material.


\begin{table}[t!]
\centering
\resizebox{0.49\textwidth}{!}{
    \renewcommand{\arraystretch}{1}
    \addtolength{\tabcolsep}{-2pt}
    \begin{tabular}{r|c|ccc|ccc}
    \hline
    Model & \begin{tabular}[x]{@{}c@{}}Wire\\IoU\end{tabular} & F1 & Precision & Recall & \begin{tabular}[x]{@{}c@{}}IoU\\(Small)\end{tabular} & \begin{tabular}[x]{@{}c@{}}IoU\\(Medium)\end{tabular} & \begin{tabular}[x]{@{}c@{}}IoU\\(Large)\end{tabular} \\\hline\hline
    Ours & 60.83 & 75.65 & 83.62 & 69.06 & 63.52 & 59.83 & 62.93 \\ \hline
    -- MinMax & 60.01 & 75.01 & 84.87 & 67.2 & 63.67 & 58.99 & 61.97 \\
    -- MaxPool & 59.86 & 74.89 & 85.25 & 66.78 & 61.45 & 59.40 & 60.76 \\
    -- Coarse & 56.92 & 72.55 & 82.91 & 64.49 & 62.83 & 57.42 & 54.47 \\ \hline
\end{tabular}
\addtolength{\tabcolsep}{2pt}
}
\vspace{-1mm}
\caption{Ablation study of our model components.
}
\label{table:component_ablation}
\vspace{-2mm}
\end{table}
\begin{table}[t!]
\centering
\resizebox{0.9\columnwidth}{!}{
    \renewcommand{\arraystretch}{1.1}
    \begin{tabular}{r|c|ccc|cc}
    \hline
    $\alpha$& \begin{tabular}[x]{@{}c@{}}Wire\\IoU\end{tabular} & F1 & Precision & Recall & \begin{tabular}[x]{@{}c@{}}Avg. Time\\(s/img)\end{tabular} &
    Speed up\\ \hline
    0.0 & 60.97 & 75.75 & 82.63 & 69.93 & 1.91 & 1$\times$ \\
    0.01 & 60.83 & 75.65 & 83.62 & 69.06 & 0.82 & 2.3$\times$ \\
    0.02 & 60.35 & 75.27 & 83.97 & 68.20 & 0.75 & 2.5$\times$ \\
    0.05 & 55.17 & 71.11 & 84.84 & 61.20 & 0.58 & 3.3$\times$ \\
    0.1 & 42.44 & 59.59 & 86.06 & 45.57 & 0.4 & 4.8$\times$ \\
    \hline
\end{tabular}
}
\vspace{-1mm}
\caption{Ablation on the threshold for refinement. At $\alpha=0.0$, all windows are passed to the fine module.}
\label{table:thresholds}
\end{table}

\begin{table}[t]\setlength{\tabcolsep}{5pt}
\setlength{\abovecaptionskip}{8pt}
\centering
\footnotesize

\begin{tabular}{r|c c c|c}
\hline
Model &PSNR$\uparrow$&LPIPS$\downarrow$&FID$\downarrow$ &Speed (s/img)\\ \hline
PatchMatch \cite{barnes2009patchmatch}&50.29 &0.0294 & 5.0403 & -\\
DeepFillv2 \cite{yu2019free} &47.01 &0.0374&8.0086 &0.009\\
CMGAN \cite{zheng2022cm} &50.07 &0.0255 &3.8286 &0.141\\
FcF \cite{jain2022keys}&48.82&0.0322&4.7848&0.048\\
LDM \cite{rombach2022high} & 45.96 & 0.0401& 10.1687 & 4.280\\
Big-LaMa \cite{suvorov2022resolution} & 49.63 & 0.0267& 4.1245 &0.034\\
Ours (LaMa-Wire) & 50.06 & 0.0259 & 3.6950 &0.034\\
\hline
\end{tabular}
\vspace{-1mm}
\caption{Quantitative results of inpainting on our synthetic wire inpainting evaluation dataset (1000 images). Our model achieves the highest perceptual quality in terms of FID, and has a balanced speed and quality.}
\label{exp:wire_inp}
\end{table}

\section{Conclusion}

In this paper, we propose a fully automated wire segmentation and removal system for high-resolution imagery. We demonstrate a segmentation method that maximally preserves sparse wire features and annotations, with a two-stage model that effectively uses global context and local details. The predicted segmentation mask is used in our tile-based wire inpainting model that has been demonstrated to produce seamless inpainting results.
We also introduce \benchmark, the first benchmark wire dataset with high-quality annotations. We hope our proposed method will provide insights into tackling semantic segmentation with high resolution image and annotation properties, and that our benchmark dataset encourage further research in wire segmentation and removal.


{\small
\bibliographystyle{ieee_fullname}
\bibliography{egbib}

\begin{thebibliography}{10}\itemsep=-1pt

\bibitem{Pixel6as49}
Pixel 6, a smarter chip for a smarter phone - google store.
\newblock \url{https://store.google.com/product/pixel_6?hl=en-US}.
\newblock (Accessed on 11/14/2021).

\bibitem{ttpla}
Rabab Abdelfattah, Xiaofeng Wang, and Song Wang.
\newblock Ttpla: An aerial-image dataset for detection and segmentation of
  transmission towers and power lines.
\newblock In {\em Proceedings of the Asian Conference on Computer Vision},
  2020.

\bibitem{barnes2009patchmatch}
Connelly Barnes, Eli Shechtman, Adam Finkelstein, and Dan~B Goldman.
\newblock Patchmatch: A randomized correspondence algorithm for structural
  image editing.
\newblock {\em ACM Trans. Graph.}, 28(3):24, 2009.

\bibitem{deeplab}
Liang-Chieh Chen, George Papandreou, Iasonas Kokkinos, Kevin Murphy, and Alan~L
  Yuille.
\newblock Deeplab: Semantic image segmentation with deep convolutional nets,
  atrous convolution, and fully connected crfs.
\newblock {\em IEEE transactions on pattern analysis and machine intelligence},
  40(4):834--848, 2017.

\bibitem{deeplabv3}
Liang-Chieh Chen, George Papandreou, Florian Schroff, and Hartwig Adam.
\newblock Rethinking atrous convolution for semantic image segmentation.
\newblock {\em arXiv preprint arXiv:1706.05587}, 2017.

\bibitem{deeplabv3p}
Liang-Chieh Chen, Yukun Zhu, George Papandreou, Florian Schroff, and Hartwig
  Adam.
\newblock Encoder-decoder with atrous separable convolution for semantic image
  segmentation.
\newblock In {\em Proceedings of the European conference on computer vision
  (ECCV)}, pages 801--818, 2018.

\bibitem{glnet}
Wuyang Chen, Ziyu Jiang, Zhangyang Wang, Kexin Cui, and Xiaoning Qian.
\newblock Collaborative global-local networks for memory-efficient segmentation
  of ultra-high resolution images.
\newblock In {\em Proceedings of the IEEE/CVF Conference on Computer Vision and
  Pattern Recognition}, pages 8924--8933, 2019.

\bibitem{cascadepsp}
Ho~Kei Cheng, Jihoon Chung, Yu-Wing Tai, and Chi-Keung Tang.
\newblock Cascadepsp: toward class-agnostic and very high-resolution
  segmentation via global and local refinement.
\newblock In {\em Proceedings of the IEEE/CVF Conference on Computer Vision and
  Pattern Recognition}, pages 8890--8899, 2020.

\bibitem{darabi2012image}
Soheil Darabi, Eli Shechtman, Connelly Barnes, Dan~B Goldman, and Pradeep Sen.
\newblock Image melding: Combining inconsistent images using patch-based
  synthesis.
\newblock {\em ACM Transactions on graphics (TOG)}, 31(4):1--10, 2012.

\bibitem{isdnet}
Shaohua Guo, Liang Liu, Zhenye Gan, Yabiao Wang, Wuhao Zhang, Chengjie Wang,
  Guannan Jiang, Wei Zhang, Ran Yi, Lizhuang Ma, et~al.
\newblock Isdnet: Integrating shallow and deep networks for efficient
  ultra-high resolution segmentation.
\newblock In {\em Proceedings of the IEEE/CVF Conference on Computer Vision and
  Pattern Recognition}, pages 4361--4370, 2022.

\bibitem{hassani2022dilated}
Ali Hassani and Humphrey Shi.
\newblock Dilated neighborhood attention transformer.
\newblock 2022.

\bibitem{hassani2022neighborhood}
Ali Hassani, Steven Walton, Jiachen Li, Shen Li, and Humphrey Shi.
\newblock Neighborhood attention transformer.
\newblock 2022.

\bibitem{resnet}
Kaiming He, Xiangyu Zhang, Shaoqing Ren, and Jian Sun.
\newblock Deep residual learning for image recognition.
\newblock In {\em Proceedings of the IEEE conference on computer vision and
  pattern recognition}, pages 770--778, 2016.

\bibitem{ccnet}
Zilong Huang, Xinggang Wang, Lichao Huang, Chang Huang, Yunchao Wei, and Wenyu
  Liu.
\newblock Ccnet: Criss-cross attention for semantic segmentation.
\newblock In {\em Proceedings of the IEEE/CVF International Conference on
  Computer Vision}, pages 603--612, 2019.

\bibitem{magnet}
Chuong Huynh, Anh~Tuan Tran, Khoa Luu, and Minh Hoai.
\newblock Progressive semantic segmentation.
\newblock In {\em Proceedings of the IEEE/CVF Conference on Computer Vision and
  Pattern Recognition}, pages 16755--16764, 2021.

\bibitem{globallocal}
Satoshi Iizuka, Edgar Simo-Serra, and Hiroshi Ishikawa.
\newblock Globally and locally consistent image completion.
\newblock {\em ACM Transactions on Graphics (ToG)}, 36(4):1--14, 2017.

\bibitem{jain2022oneformer}
Jitesh Jain, Jiachen Li, MangTik Chiu, Ali Hassani, Nikita Orlov, and Humphrey
  Shi.
\newblock {OneFormer: One Transformer to Rule Universal Image Segmentation}.
\newblock 2023.

\bibitem{jain2021semask}
Jitesh Jain, Anukriti Singh, Nikita Orlov, Zilong Huang, Jiachen Li, Steven
  Walton, and Humphrey Shi.
\newblock Semask: Semantically masking transformer backbones for effective
  semantic segmentation.
\newblock {\em arXiv}, 2021.

\bibitem{jain2022keys}
Jitesh Jain, Yuqian Zhou, Ning Yu, and Humphrey Shi.
\newblock Keys to better image inpainting: Structure and texture go hand in
  hand.
\newblock {\em arXiv preprint arXiv:2208.03382}, 2022.

\bibitem{swiftlane}
Oshada Jayasinghe, Damith Anhettigama, Sahan Hemachandra, Shenali Kariyawasam,
  Ranga Rodrigo, and Peshala Jayasekara.
\newblock Swiftlane: Towards fast and efficient lane detection.
\newblock {\em arXiv preprint arXiv:2110.11779}, 2021.

\bibitem{karras2020analyzing}
Tero Karras, Samuli Laine, Miika Aittala, Janne Hellsten, Jaakko Lehtinen, and
  Timo Aila.
\newblock Analyzing and improving the image quality of stylegan.
\newblock In {\em Proceedings of the IEEE/CVF conference on computer vision and
  pattern recognition}, pages 8110--8119, 2020.

\bibitem{kaspar2015self}
Alexandre Kaspar, Boris Neubert, Dani Lischinski, Mark Pauly, and Johannes
  Kopf.
\newblock Self tuning texture optimization.
\newblock In {\em Computer Graphics Forum}, volume~34, pages 349--359. Wiley
  Online Library, 2015.

\bibitem{cable_inst}
Bo Li, Cheng Chen, Shiwen Dong, and Junfeng Qiao.
\newblock Transmission line detection in aerial images: An instance
  segmentation approach based on multitask neural networks.
\newblock {\em Signal Processing: Image Communication}, 96:116278, 2021.

\bibitem{focal}
Tsung-Yi Lin, Priya Goyal, Ross Girshick, Kaiming He, and Piotr Doll{\'a}r.
\newblock Focal loss for dense object detection.
\newblock In {\em Proceedings of the IEEE international conference on computer
  vision}, pages 2980--2988, 2017.

\bibitem{partialconv}
Guilin Liu, Fitsum~A Reda, Kevin~J Shih, Ting-Chun Wang, Andrew Tao, and Bryan
  Catanzaro.
\newblock Image inpainting for irregular holes using partial convolutions.
\newblock In {\em Proceedings of the European conference on computer vision
  (ECCV)}, pages 85--100, 2018.

\bibitem{swin}
Ze Liu, Yutong Lin, Yue Cao, Han Hu, Yixuan Wei, Zheng Zhang, Stephen Lin, and
  Baining Guo.
\newblock Swin transformer: Hierarchical vision transformer using shifted
  windows.
\newblock {\em arXiv preprint arXiv:2103.14030}, 2021.

\bibitem{adamw}
Ilya Loshchilov and Frank Hutter.
\newblock Decoupled weight decay regularization.
\newblock {\em arXiv preprint arXiv:1711.05101}, 2017.

\bibitem{lsnet}
Van~Nhan Nguyen, Robert Jenssen, and Davide Roverso.
\newblock Ls-net: Fast single-shot line-segment detector.
\newblock {\em arXiv preprint arXiv:1912.09532}, 2019.

\bibitem{contextencoder}
Deepak Pathak, Philipp Krahenbuhl, Jeff Donahue, Trevor Darrell, and Alexei~A
  Efros.
\newblock Context encoders: Feature learning by inpainting.
\newblock In {\em Proceedings of the IEEE conference on computer vision and
  pattern recognition}, pages 2536--2544, 2016.

\bibitem{dalle}
Aditya Ramesh, Prafulla Dhariwal, Alex Nichol, Casey Chu, and Mark Chen.
\newblock Hierarchical text-conditional image generation with clip latents.
\newblock {\em arXiv preprint arXiv:2204.06125}, 2022.

\bibitem{dpt}
Ren{\'e} Ranftl, Alexey Bochkovskiy, and Vladlen Koltun.
\newblock Vision transformers for dense prediction.
\newblock In {\em Proceedings of the IEEE/CVF International Conference on
  Computer Vision}, pages 12179--12188, 2021.

\bibitem{rombach2022high}
Robin Rombach, Andreas Blattmann, Dominik Lorenz, Patrick Esser, and Bj{\"o}rn
  Ommer.
\newblock High-resolution image synthesis with latent diffusion models.
\newblock In {\em Proceedings of the IEEE/CVF Conference on Computer Vision and
  Pattern Recognition}, pages 10684--10695, 2022.

\bibitem{ohem}
Abhinav Shrivastava, Abhinav Gupta, and Ross Girshick.
\newblock Training region-based object detectors with online hard example
  mining.
\newblock In {\em Proceedings of the IEEE conference on computer vision and
  pattern recognition}, pages 761--769, 2016.

\bibitem{structurelane}
Jinming Su, Chao Chen, Ke Zhang, Junfeng Luo, Xiaoming Wei, and Xiaolin Wei.
\newblock Structure guided lane detection.
\newblock {\em arXiv preprint arXiv:2105.05403}, 2021.

\bibitem{suvorov2022resolution}
Roman Suvorov, Elizaveta Logacheva, Anton Mashikhin, Anastasia Remizova,
  Arsenii Ashukha, Aleksei Silvestrov, Naejin Kong, Harshith Goka, Kiwoong
  Park, and Victor Lempitsky.
\newblock Resolution-robust large mask inpainting with fourier convolutions.
\newblock In {\em Proceedings of the IEEE/CVF Winter Conference on Applications
  of Computer Vision}, pages 2149--2159, 2022.

\bibitem{lanedet}
Lucas Tabelini, Rodrigo Berriel, Thiago~M Paixao, Claudine Badue, Alberto~F
  De~Souza, and Thiago Oliveira-Santos.
\newblock Keep your eyes on the lane: Real-time attention-guided lane
  detection.
\newblock In {\em Proceedings of the IEEE/CVF Conference on Computer Vision and
  Pattern Recognition}, pages 294--302, 2021.

\bibitem{attention}
Ashish Vaswani, Noam Shazeer, Niki Parmar, Jakob Uszkoreit, Llion Jones,
  Aidan~N Gomez, {\L}ukasz Kaiser, and Illia Polosukhin.
\newblock Attention is all you need.
\newblock In {\em Advances in neural information processing systems}, pages
  5998--6008, 2017.

\bibitem{wexler2007space}
Yonatan Wexler, Eli Shechtman, and Michal Irani.
\newblock Space-time completion of video.
\newblock {\em IEEE Transactions on pattern analysis and machine intelligence},
  29(3):463--476, 2007.

\bibitem{segformer}
Enze Xie, Wenhai Wang, Zhiding Yu, Anima Anandkumar, Jose~M Alvarez, and Ping
  Luo.
\newblock Segformer: Simple and efficient design for semantic segmentation with
  transformers.
\newblock {\em arXiv preprint arXiv:2105.15203}, 2021.

\bibitem{xu2022image}
Xingqian Xu, Shant Navasardyan, Vahram Tadevosyan, Andranik Sargsyan, Yadong
  Mu, and Humphrey Shi.
\newblock Image completion with heterogeneously filtered spectral hints.
\newblock In {\em WACV}, 2023.

\bibitem{powerlinedataset}
{\"O}mer~Emre Yetgin, {\"O}mer~Nezih Gerek, and {\"O}mer Nezih.
\newblock Power image dataset (infrared-ir and visible light-vl).
\newblock {\em Mendeley Data}, 8, 2017.

\bibitem{hifill}
Zili Yi, Qiang Tang, Shekoofeh Azizi, Daesik Jang, and Zhan Xu.
\newblock Contextual residual aggregation for ultra high-resolution image
  inpainting.
\newblock In {\em Proceedings of the IEEE/CVF Conference on Computer Vision and
  Pattern Recognition}, pages 7508--7517, 2020.

\bibitem{contextual}
Jiahui Yu, Zhe Lin, Jimei Yang, Xiaohui Shen, Xin Lu, and Thomas~S Huang.
\newblock Generative image inpainting with contextual attention.
\newblock In {\em Proceedings of the IEEE conference on computer vision and
  pattern recognition}, pages 5505--5514, 2018.

\bibitem{yu2019free}
Jiahui Yu, Zhe Lin, Jimei Yang, Xiaohui Shen, Xin Lu, and Thomas~S Huang.
\newblock Free-form image inpainting with gated convolution.
\newblock In {\em Proceedings of the IEEE/CVF international conference on
  computer vision}, pages 4471--4480, 2019.

\bibitem{zeng2020high}
Yu Zeng, Zhe Lin, Jimei Yang, Jianming Zhang, Eli Shechtman, and Huchuan Lu.
\newblock High-resolution image inpainting with iterative confidence feedback
  and guided upsampling.
\newblock In {\em European conference on computer vision}, pages 1--17.
  Springer, 2020.

\bibitem{pldu}
Heng Zhang, Wen Yang, Huai Yu, Haijian Zhang, and Gui-Song Xia.
\newblock Detecting power lines in uav images with convolutional features and
  structured constraints.
\newblock {\em Remote Sensing}, 11(11):1342, 2019.

\bibitem{supercaf}
Lingzhi Zhang, Connelly Barnes, Kevin Wampler, Sohrab Amirghodsi, Eli
  Shechtman, Zhe Lin, and Jianbo Shi.
\newblock Inpainting at modern camera resolution by guided patchmatch with
  auto-curation.
\newblock In {\em European Conference on Computer Vision}, pages 51--67.
  Springer, 2022.

\bibitem{pspnet}
Hengshuang Zhao, Jianping Shi, Xiaojuan Qi, Xiaogang Wang, and Jiaya Jia.
\newblock Pyramid scene parsing network.
\newblock In {\em Proceedings of the IEEE conference on computer vision and
  pattern recognition}, pages 2881--2890, 2017.

\bibitem{comodgan}
Shengyu Zhao, Jonathan Cui, Yilun Sheng, Yue Dong, Xiao Liang, Eric~I Chang,
  and Yan Xu.
\newblock Large scale image completion via co-modulated generative adversarial
  networks.
\newblock {\em arXiv preprint arXiv:2103.10428}, 2021.

\bibitem{zheng2022cm}
Haitian Zheng, Zhe Lin, Jingwan Lu, Scott Cohen, Eli Shechtman, Connelly
  Barnes, Jianming Zhang, Ning Xu, Sohrab Amirghodsi, and Jiebo Luo.
\newblock Cm-gan: Image inpainting with cascaded modulation gan and
  object-aware training.
\newblock {\em arXiv preprint arXiv:2203.11947}, 2022.

\bibitem{setr}
Sixiao Zheng, Jiachen Lu, Hengshuang Zhao, Xiatian Zhu, Zekun Luo, Yabiao Wang,
  Yanwei Fu, Jianfeng Feng, Tao Xiang, Philip~HS Torr, et~al.
\newblock Rethinking semantic segmentation from a sequence-to-sequence
  perspective with transformers.
\newblock In {\em Proceedings of the IEEE/CVF Conference on Computer Vision and
  Pattern Recognition}, pages 6881--6890, 2021.

\bibitem{zhou2017places}
Bolei Zhou, Agata Lapedriza, Aditya Khosla, Aude Oliva, and Antonio Torralba.
\newblock Places: A 10 million image database for scene recognition.
\newblock {\em IEEE Transactions on Pattern Analysis and Machine Intelligence},
  2017.

\end{thebibliography}


\begin{thebibliography}{1}\itemsep=-1pt

\bibitem{cascadepsp}
Ho~Kei Cheng, Jihoon Chung, Yu-Wing Tai, and Chi-Keung Tang.
\newblock Cascadepsp: toward class-agnostic and very high-resolution
  segmentation via global and local refinement.
\newblock In {\em Proceedings of the IEEE/CVF Conference on Computer Vision and
  Pattern Recognition}, pages 8890--8899, 2020.

\bibitem{isdnet}
Shaohua Guo, Liang Liu, Zhenye Gan, Yabiao Wang, Wuhao Zhang, Chengjie Wang,
  Guannan Jiang, Wei Zhang, Ran Yi, Lizhuang Ma, et~al.
\newblock Isdnet: Integrating shallow and deep networks for efficient
  ultra-high resolution segmentation.
\newblock In {\em Proceedings of the IEEE/CVF Conference on Computer Vision and
  Pattern Recognition}, pages 4361--4370, 2022.

\bibitem{magnet}
Chuong Huynh, Anh~Tuan Tran, Khoa Luu, and Minh Hoai.
\newblock Progressive semantic segmentation.
\newblock In {\em Proceedings of the IEEE/CVF Conference on Computer Vision and
  Pattern Recognition}, pages 16755--16764, 2021.

\end{thebibliography}
}

\end{document}



\maketitle
\thispagestyle{empty}

\vspace{-1mm}
\section{Comparison with Pixel 6}
\begin{figure}[h!]
    \setlength{\abovecaptionskip}{1mm}
    \centering
    \captionsetup{type=figure}
    \includegraphics[width=0.43\textwidth]{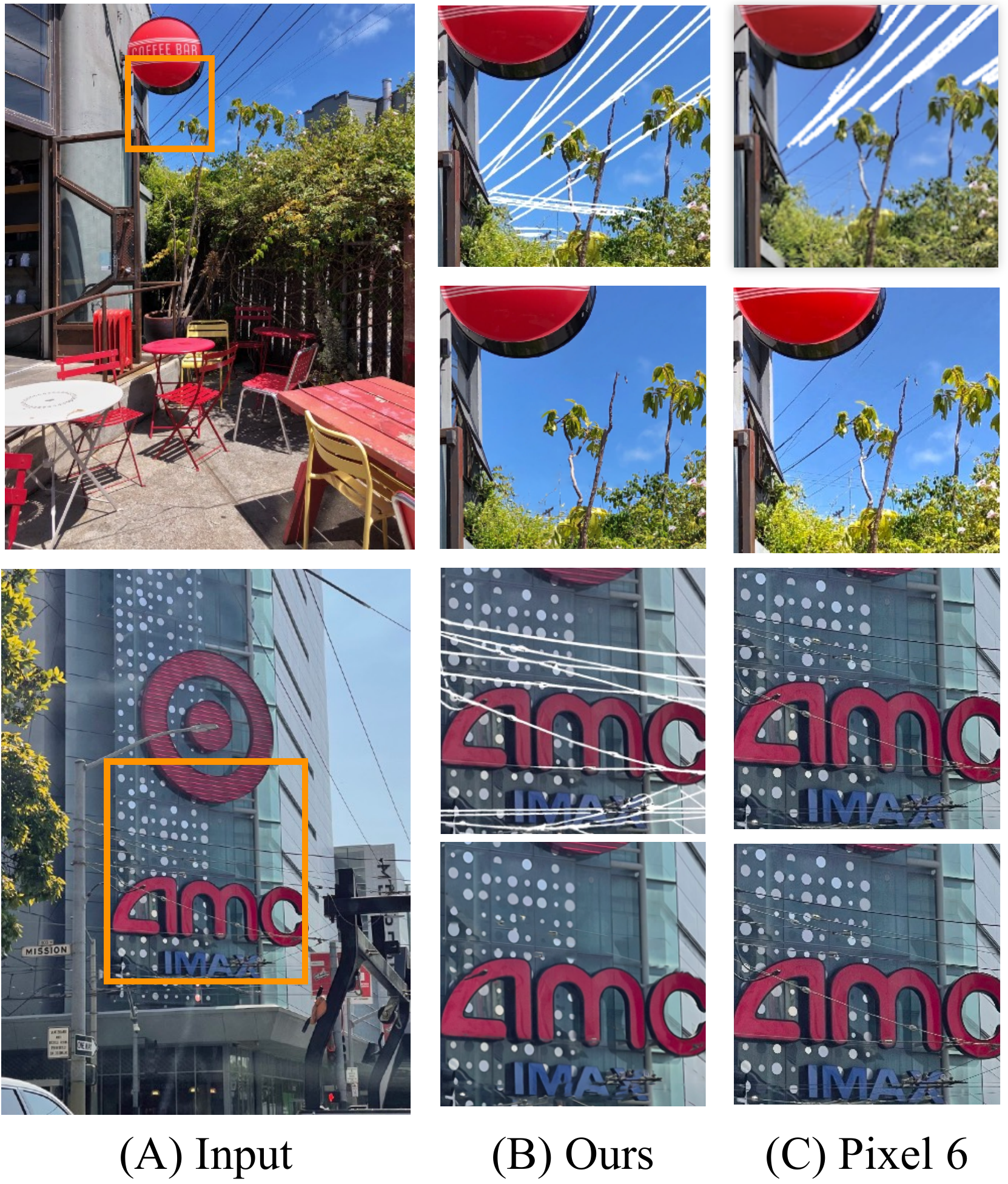}
    \vspace{-1mm}
    \captionof{figure}{\textbf{Comparison with Pixel 6}. Our model can pick up hardly visible wires that even in complicated backgrounds}
    \label{fig:pixel6}
\end{figure}

\vspace{-1mm}
We show a visual comparison between our model and Pixel 6's ``Magic Eraser'' feature in Figure~\ref{fig:pixel6}. Without manual intervention, Google Pixel 6's ``Magic Eraser'' performs well on wires with clean background, but suffers from thin wires that are hardly visible ((A) upper), and also on wires with complicated background ((A) lower). We also pass our segmentation mask to our wire inpainting model to acquire the wire removal result, as shown in the lower image of (B).

\vspace{-1mm}
\section{Failure cases}
\begin{figure}[h!]
    \setlength{\abovecaptionskip}{1mm}
    \centering
    \includegraphics[width=0.9\linewidth]{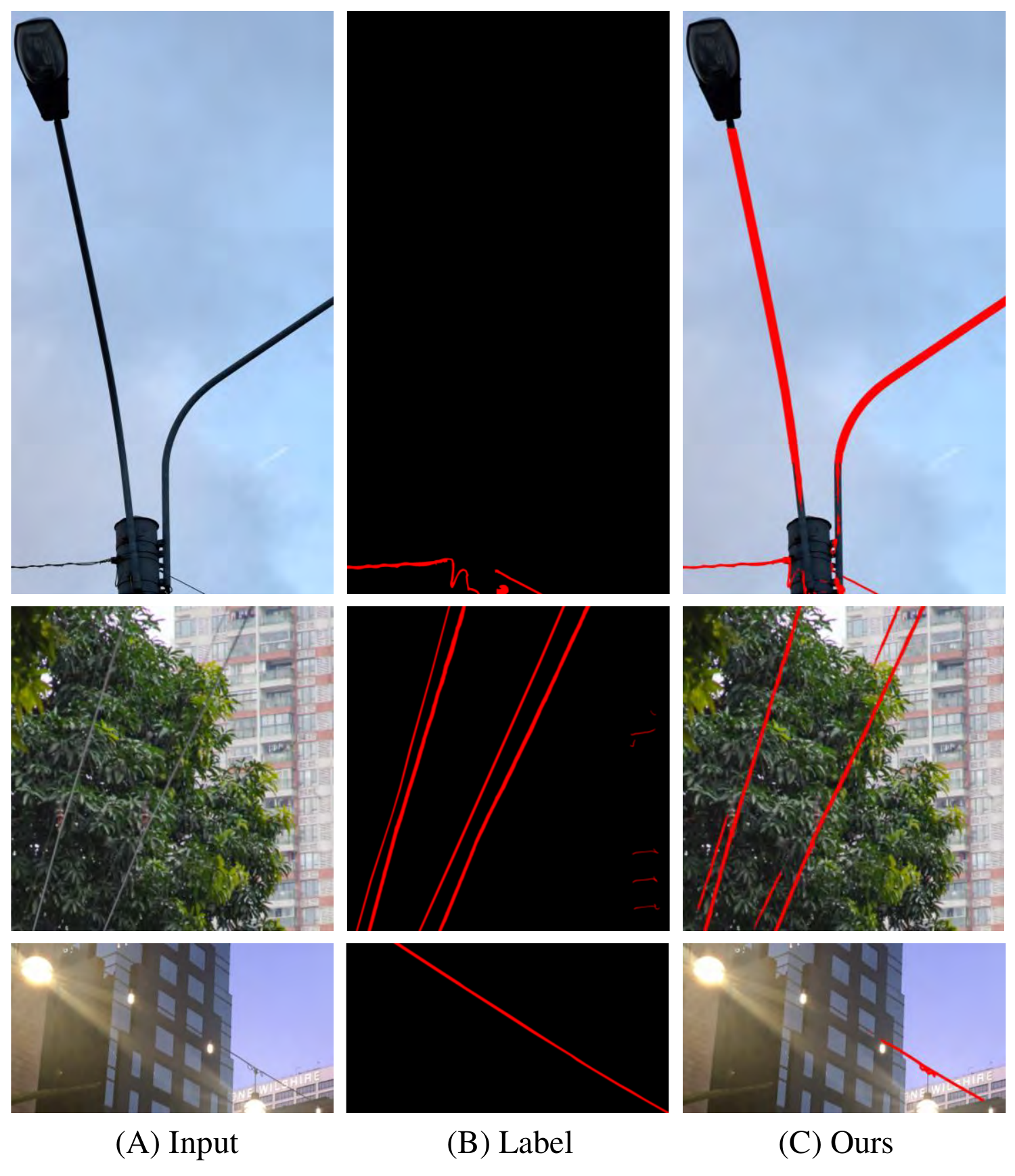}
    \caption{\textbf{Failure cases}. In some challenging cases, our model fails to predict accurate masks. Zoom in to see detailed wire masks in ground truth and prediction.}
    \label{fig:new_failures}
    \vspace{-10mm}
\end{figure}

\vspace{-1mm}
We show some challenging cases where our model fails to predict accurate wire masks in Figure~\ref{fig:new_failures}. These include regions that are very similar to wires (top row), severe background blending (middle row) and extreme lighting conditions (bottom row).

\vspace{-1mm}
\section{Panorama}
\begin{figure*}[t!]
\centering
\includegraphics[width=1.0\textwidth]{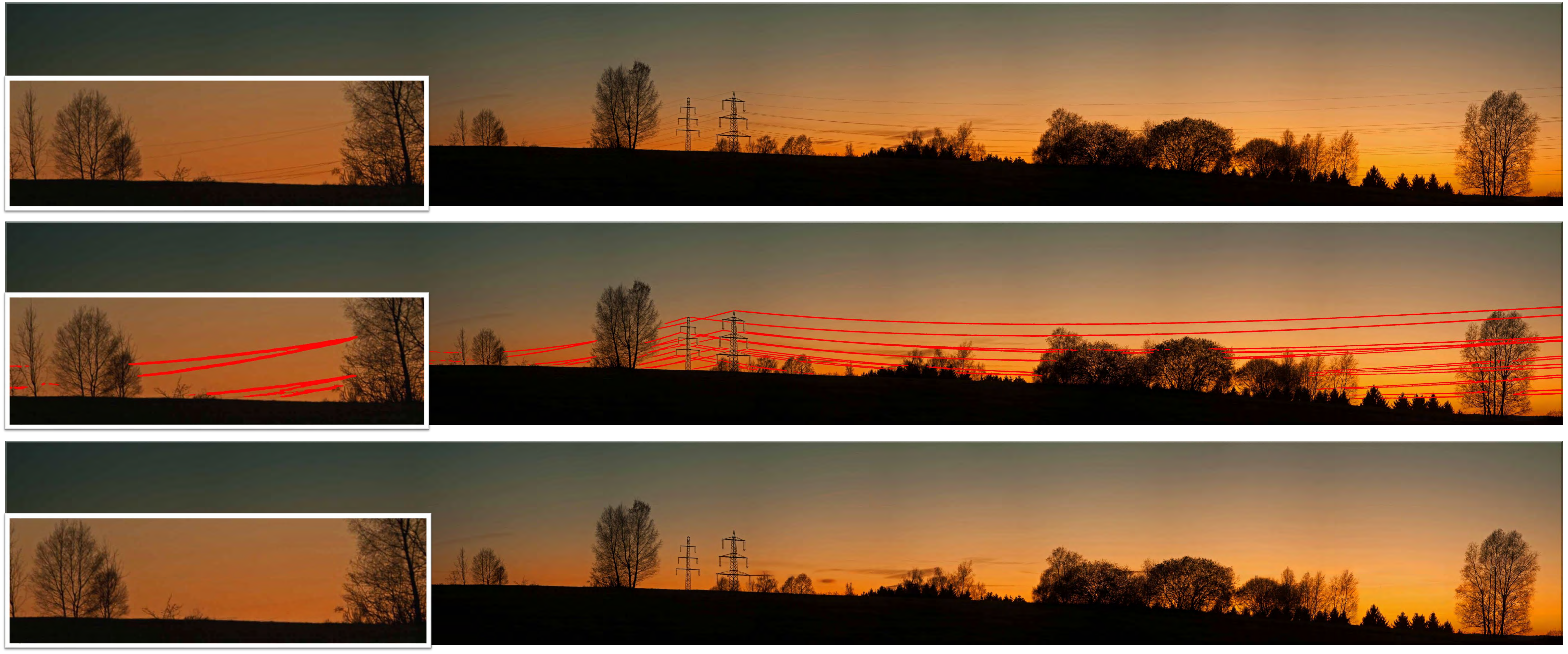}
\caption{\textbf{Segmentation and inpainting result for a panoramic image.} Our model is scalable to very large images with very thin wires.}
\label{fig:new_panorama}
\vspace{-3mm}
\end{figure*}

\vspace{-1mm}
Our two-stage model leverages the sparsity of wires in natural images, and efficiently generalizes to ultra-high resolution images such as panoramas. We show one panoramic image of $11$K by $1.5$K resolution in Figure~\ref{fig:new_panorama}. Note that our method produces high-quality wire segmentation that covers wires that are almost invisible. As a result, our proposed wire removal step can effectively remove these regions.


\section{Segmentation and inpainting visualizations}
\begin{figure*}[t!]
    \setlength{\abovecaptionskip}{1mm}
    \centering
    \includegraphics[width=\textwidth]{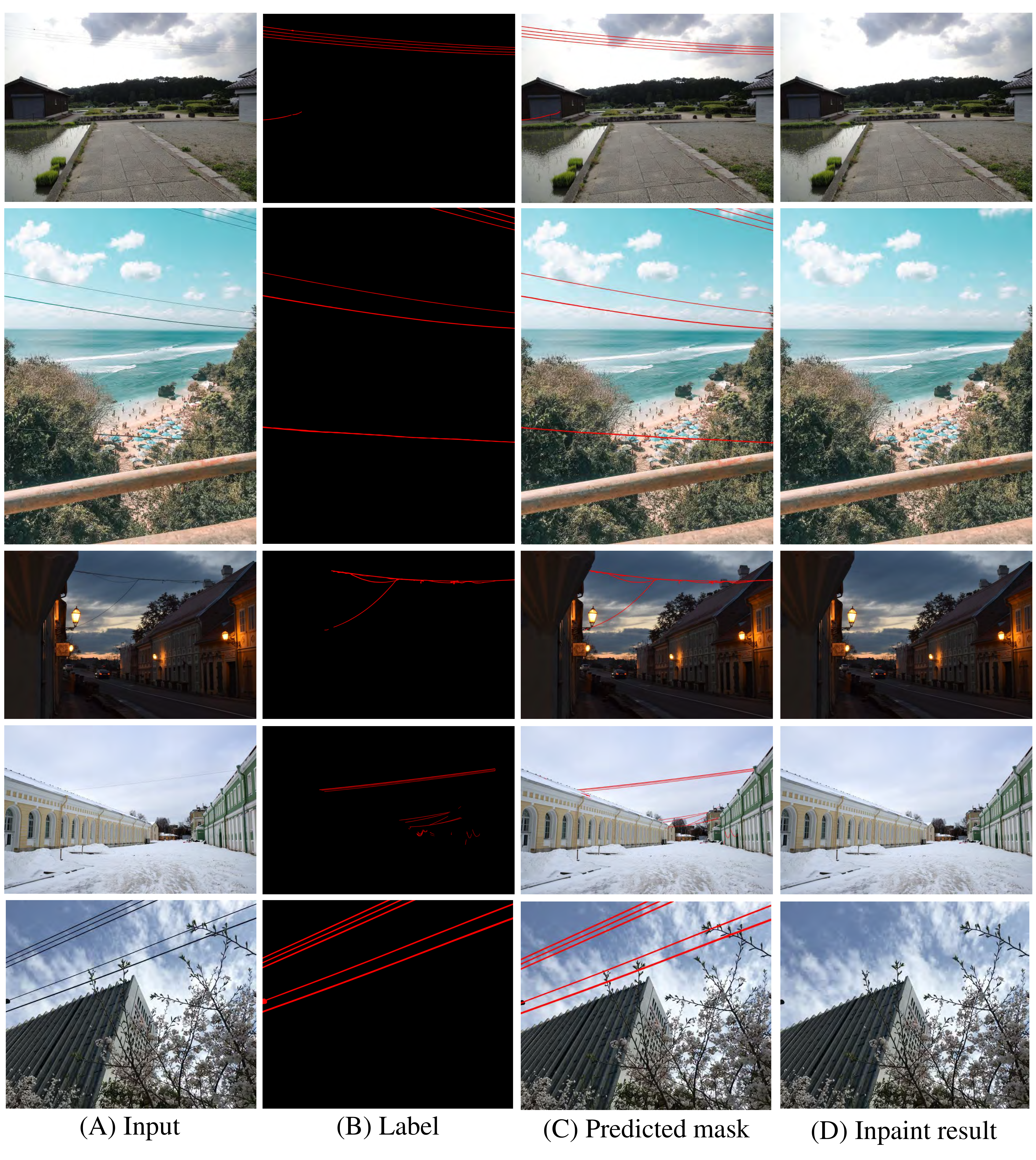}
    \caption{\textbf{Segmentation and inpainting visualizations}. Our model can handle several challenging scenes, including strongly backlit (top row), background with complex texture (2nd row), low light (3rd row), and barely visible wires (4th row)}
    \label{fig:additional_visualizations}
\end{figure*}

\vspace{-1mm}
We show our wire segmentation and inpainting results in several common photography scenes as well as in some challenging cases in Figure~\ref{fig:additional_visualizations}. We provide more visualizations of wire segmentation and subsequent inpainting results. Our model successfully handles numerous challenging scenarios, including strong backlit (top row), complex background texture (2nd row), low light (3rd row), and barely visible wires (4th row). A typical use case is shown in the last row. 

\section{Experiments on other datasets}
\vspace{-1mm}
Most existing wire-like datasets either are at low resolutions or are for specific purposes (e.g., aerial imaging) and thus do not contain the scene diversity like WireSegHR does. The suggested TTPLA~[2] dataset shares the Power Lines class with our dataset, although it only contains aerial images. Table~\ref{ttpla_exp} shows evaluation results of the TTPLA dataset on our model and also our WireSegHR dataset on the TTPLA model.

\begin{table}[h!]
\resizebox{\linewidth}{!}{
    \centering
    \begin{tabular}{c|c|c}
        Dataset & Model & IoU (\%) \\\hline\hline
        \multirow{3}{*}{TTPLA (Power Line only)} & TTPLA (ResNet-50, 700$\times$ 700) & 18.9 \\
        & Ours (ResNet-50) & 33.1 \\
        & Ours (MiT-b2) & 42.7 \\\hline
        \multirow{3}{*}{WireSegHR} & TTPLA (ResNet-50, 700$\times$ 700) & 3.5 \\
        & Ours (ResNet-50) & 47.8 \\
        & Ours (MiT-b2) & 60.8\\\hline
    \end{tabular}}
\caption{Comparison with TTPLA.}
\vspace{-5mm}
\label{ttpla_exp}
\end{table}

TTPLA is trained on fixed resolution (700 $\times$ 700) and takes in the entire image for inference, which requires significant downsampling of our test set.
As a result, the quality of thin wires deteriorates in both the image and the label. Our model drops in performance on the TTPLA dataset due to different annotation definitions: we annotate all wire-like objects while TTPLA only annotates power lines.

\vspace{-1mm}
\section{Additional training details}
\vspace{-1mm}
\paragraph{CascadePSP~\cite{cascadepsp}}
We follow the default training steps provided by the CascadePSP code\footnote{\label{note1}\href{https://github.com/hkchengrex/CascadePSP}{https://github.com/hkchengrex/CascadePSP}}. During training, we sample patches in the image that contain at least 1\% of wire pixels. During inference, we feed the predictions of the global DeepLabv3+ to the pretrained/retrained CascadePSP model to get the refined wire mask. In both cases, we follow the default inference code\footnoteref{note1} to obtain the final mask.
\vspace{-3mm}
\paragraph{MagNet~\cite{magnet}} MagNet\footnote{\href{https://github.com/VinAIResearch/MagNet}{https://github.com/VinAIResearch/MagNet}} obtains the initial mask predictions from a single backbone trained on all refinement scales. For a fair comparison, we adopt a 2-scale setting of MagNet, similar to our two-stage model, where the image is downsampled to $1024\times 1024$ in the global scale, and is kept at the original resolution in the local scale. To this end, we train a single DeepLabv3+ model by either downsampling the sample image to $1024\times 1024$ or randomly cropping $1024\times 1024$ patches at the original resolution. The sampled patches contain at least 1\% of wire pixels. We then train the refinement module based on the predictions from the DeepLabv3+ model, following the default setting. Inference is kept the same as the original MagNet model.

\vspace{-2mm}
\paragraph{ISDNet~\cite{isdnet}}
ISDNet\footnote{\href{https://github.com/cedricgsh/ISDNet}{https://github.com/cedricgsh/ISDNet}} performs inference on the entire image without sliding window. As a result, during training, we resize all images to $5000\times 5000$ and randomly crop $2500\times 2500$ windows, such that the input images can fit inside the GPUs. Sampled patches should contain 1\% wire pixels. During inference, all images are resized to $5000\times 5000$. We observe that this yields better results than if we keep images below $5000\times 5000$ at their original sizes.

{\small
\bibliographystyle{ieee_fullname}
\bibliography{egbib}
}